\documentclass{article}

\usepackage{microtype}
\usepackage{graphicx}
\usepackage{subfigure}
\usepackage{booktabs} 
\usepackage{multirow}
\usepackage{makecell}

\usepackage{hyperref}

\usepackage[accepted]{icml2025}

\usepackage{amsmath}
\usepackage{amssymb}
\usepackage{mathtools}
\usepackage{amsthm}

\usepackage[capitalize,noabbrev]{cleveref}

\theoremstyle{plain}

\theoremstyle{definition}

\theoremstyle{remark}

\usepackage[textsize=tiny]{todonotes}

\begin{document}

\twocolumn[
\icmltitle{UDiTQC: U-Net-Style Diffusion Transformer for Quantum Circuit Synthesis}




\begin{icmlauthorlist}
\icmlauthor{Zhiwei Chen}{yyy}
\icmlauthor{Hao Tang}{comp}
\end{icmlauthorlist}

\icmlaffiliation{yyy}{School of Integrated Circuit Science and Engineering, Beihang University, Beijing, China}
\icmlaffiliation{comp}{School of Computer Science, Peking University,  Beijing, China}

\icmlcorrespondingauthor{Hao Tang}{haotang@pku.edu.cn}

\icmlkeywords{Diffusion model, Diffusion Transformer, U-Net style, Quantum circuit synthesis}

\vskip 0.3in
]


\printAffiliationsAndNotice{}  

\begin{abstract}


Quantum computing is a transformative technology with wide-ranging applications, and efficient quantum circuit generation is crucial for unlocking its full potential. Current diffusion model approaches based on U-Net architectures, while promising, encounter challenges related to computational efficiency and modeling global context. To address these issues, we propose UDiT, a novel U-Net-style Diffusion Transformer architecture, which combines U-Net’s strengths in multi-scale feature extraction with the Transformer’s ability to model global context. Building upon this foundation, we introduce UDiTQC, an extension specifically designed for quantum circuit generation. We demonstrate the framework’s effectiveness on two tasks: entanglement generation and unitary compilation, where UDiTQC consistently outperforms existing methods. Additionally, our framework supports tasks such as masking and editing circuits to meet specific physical property requirements. This dual advancement, improving quantum circuit synthesis and refining generative model architectures, marks a significant milestone in the convergence of quantum computing and machine learning research.


\end{abstract}

\section{Introduction}


Quantum computing is widely recognized as a transformative technology with great potential in fields ranging from fundamental physics research \cite{feynman2018simulating} to applications in machine learning \cite{cerezo2022challenges} and optimization \cite{farhi2014quantum}. As quantum processor technology continues to progress, the need for accurate and efficient quantum circuits has become increasingly critical. Given the constraints on available resources and gates, optimizing quantum circuit structures is essential. This requires advanced research and expertise to identify the most effective designs.


Significant progress in quantum circuit synthesis has been driven by artificial intelligence-based generative methods. These include approaches for quantum state preparation \cite{arrazola2019machine,bolens2021reinforcement}, circuit optimization \cite{fosel2021quantum,ostaszewski2021reinforcement}, and unitary compilation \cite{zhang2020topological, moro2021quantum}. Among these methods, denoising diffusion models (DMs), state-of-the-art machine learning generative techniques, have demonstrated their capability to generate quantum circuits tailored to specific requirements \cite{furrutter2024quantum}. By providing a prompt corresponding to a particular task or class, diffusion models can produce desired quantum circuit structures, making them broadly applicable to quantum computer design and optimization challenges.


Despite their success, existing approaches \cite{furrutter2024quantum} that rely on U-Net architectures for diffusion models in quantum circuit generation face several limitations. These include significant computational demands, sensitivity to data distribution, and insufficient integration of global context information. The Diffusion Transformer (DiT) \cite{peebles2023scalable} addresses these issues by embedding a full Transformer network within the diffusion model. This approach combines the global modeling strengths of Transformers with the stepwise generative capabilities of diffusion models. Replacing traditional U-Net architectures \cite{ho2020denoising} with Transformer-based frameworks improves the ability to capture long-range data dependencies, thereby enhancing generation quality. Additionally, DiT’s innovative use of time-step embeddings with input sequences offers remarkable flexibility for conditional generation and multi-modal tasks, yielding state-of-the-art results in high-resolution image generation and superior scalability in spatial generation tasks.


To further enhance quantum circuit generation, we propose a novel, practical DiT framework that improves upon conventional U-Net architectures by delivering greater efficiency and accuracy. Our approach retains U-Net’s core strengths, particularly its encoder-decoder structure and multi-scale information handling through skip connections. Building on these advantages, we introduce the U-Net-style Diffusion Transformer (UDiT) architecture, which integrates the complementary strengths of both architectures while addressing their respective weaknesses. This combined approach provides improved efficiency, accuracy, and robustness, making it better suited to handle complex practical scenarios and enabling broader applications.


To validate our approach, we apply the UDiT architecture to quantum circuit design. Our proposed framework, UDiTQC, outperforms the baseline GenQC method \cite{furrutter2024quantum} in generating quantum circuits, achieving higher accuracy across a range of metrics. UDiTQC successfully produces circuits for various qubit configurations, generating designs with diverse degrees of entanglement and maintaining precise control over physical properties and constraints. The framework also excels at targeted circuit editing and masked circuit generation, meeting specific physical constraints with exceptional flexibility. Furthermore, we demonstrate the practical utility of the UDiT model by training it for unitary compilation with limited gate sets, showcasing its applicability under real-world quantum computing constraints. By achieving enhanced precision across diverse qubit configurations and maintaining operational flexibility, our architecture not only advances quantum circuit design methodologies but also establishes a new paradigm in diffusion model architectures.

\section{Preliminaries}

\subsection{Quantum Synthesis}
The automated design of quantum experiments, particularly for generating novel entangled states, has emerged as a transformative advancement in quantum physics.
Within this domain, unitary compilation represents a fundamental challenge that directly impacts the practical implementation of quantum algorithms.
Recent approaches have made significant strides in addressing these challenges. 
The GenQC framework \cite{furrutter2024quantum} introduces an innovative solution by leveraging conditional diffusion models to generate target quantum circuits through specific prompts.
Additionally, reinforcement learning techniques have shown effectiveness in optimizing gate placement within quantum circuits, particularly for specialized tasks such as unitary synthesis \cite{rietsch2024unitary}.
These methodologies not only advance the field of quantum circuit design, but also establish new paradigms for quantum information processing and experimental physics, offering scalable solutions for increasingly complex quantum systems.

\subsection{Diffusion Models}
Diffusion models present a powerful class of generative models that create data by progressively adding noise and then reversing the process to recover the original structure. 
With their high flexibility, exceptional generative quality, and robust stability, 
diffusion models have found widespread applications in scientific fields such as image generation and editing \cite{dhariwal2021diffusion}, audio synthesis \cite{kong2020diffwave}, molecular structure design \cite{hoogeboom2022equivariant}.

The denoising diffusion probabilistic model (DDPM) \cite{ho2020denoising} established a notable advancement with its foundational framework, while parallel innovations in score-based generative models \cite{song2019generative} and stochastic differential equation (SDE)-based approaches \cite{song2020score} have further advanced the state of the art, particularly in image synthesis. 
Significant improvements in generation quality and computational efficiency have been achieved through enhanced sampling strategies, such as DDIM \cite{song2020denoising} and EDM \cite{karras2022elucidating}, alongside classifier-free guidance techniques \cite{ho2022classifier}.
Moreover, the introduction of latent diffusion models \cite{rombach2022high} marked another significant advance, incorporating data compression principles to optimize efficiency.

Although early implementations predominantly relied on U-Net architectures \cite{dhariwal2021diffusion}, the computational demands of their attention mechanisms have motivated the exploration of alternative approaches. 
Vision Transformers (ViT) \cite{dosovitskiy2020image} have emerged as a compelling alternative that offers superior scalability and enhanced modeling of long-range dependencies. 
These architectural innovations have significantly expanded the practical applications of diffusion models, pushing the boundaries of generative AI capabilities.

\subsection{Diffusion Transformer}

Diffusion Transformers (DiT) \cite{peebles2023scalable} represent a paradigm shift in diffusion models by introducing the Transformer architecture as an alternative to the conventional U-Net backbone. 
This architectural innovation harnesses the Transformer's inherent capability to model global dependencies, resulting in enhanced generation quality and increased adaptability for conditional and multi-modal generation tasks.

Building upon the foundational DiT framework, recent research has further refined its training and architectural design. 
For instance, FiT \cite{lu2024fit} and VisionLLaMA \cite{chu2024visionllama} incorporate advanced techniques inspired by large language models (LLMs), such as RoPE2D and SwishGLU, to enhance the capabilities of Diffusion Transformers. 
Such progressive developments demonstrate the framework's extensibility and scalability, positioning DiT as a state-of-the-art methodology in generative modeling across a diverse spectrum of applications.

\begin{figure*}[ht]
\centering
\centerline{\includegraphics[width=0.95\textwidth]{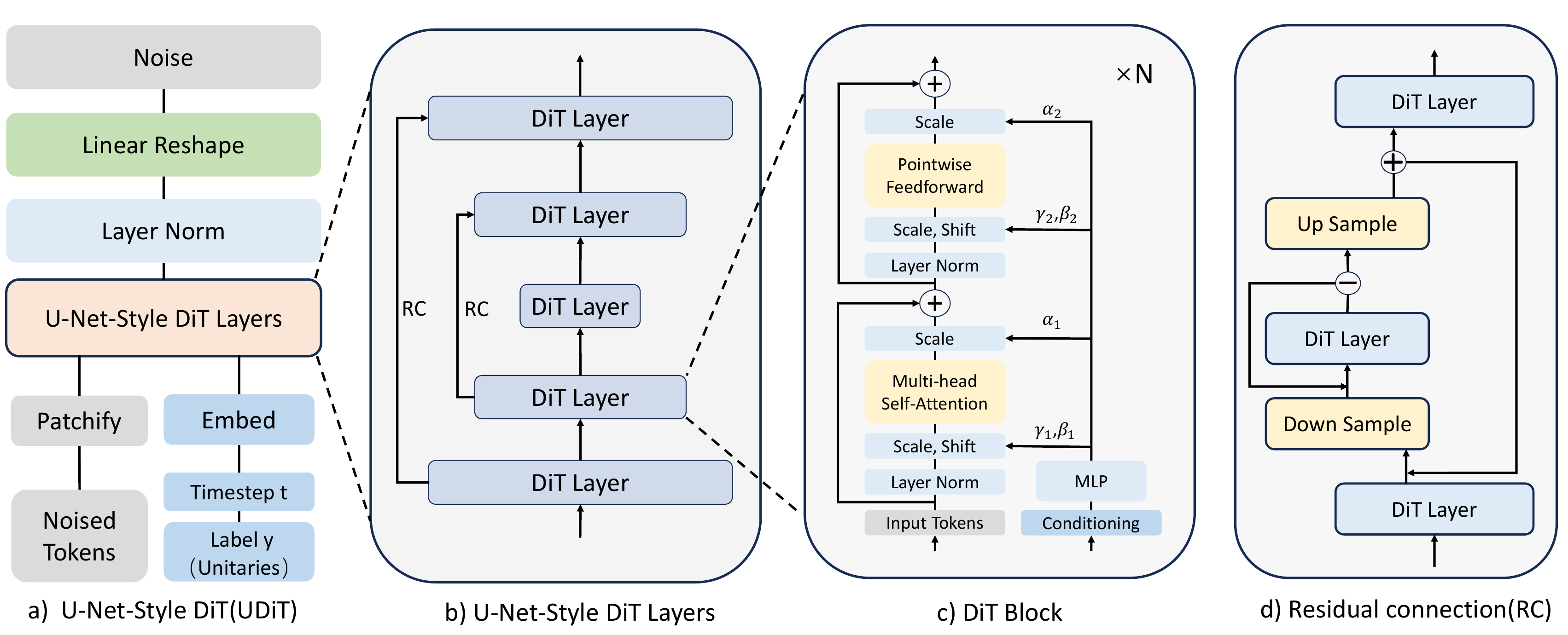}}
\caption{\textbf{The U-Net-Style Diffusion Transformer (UDiT) architecture: (a)} The overall structure of UDiT is similar to DiT, but replaces the connections between multiple DiT blocks with U-Net-style layers. \textbf{(b)} The specific form of the U-Net-style DiT layers. \textbf{(c)} Details of the DiT block, where each DiT layer is composed of N DiT blocks. \textbf{(d)} Residual connections between DiT layers.}
\label{udit}
\vspace{-0.4cm}
\end{figure*}

\section{Methodology}
Our objective is to develop more efficient diffusion model architectures. To this end, we first propose the U-Net-style Diffusion Transformer architecture (UDiT), which combines U-Net’s strengths in multi-scale feature extraction with the Transformer's ability to model global context. Building on this, we introduce UDiTQC, a framework designed for quantum circuit generation.

\subsection{The Proposed UDiT}

\subsubsection{Diffusion Models}
Before introducing our architecture, we briefly review the process of denoising diffusion probabilistic models (DDPMs).
It starts from a stochastic process where an initial sample $x_0$ is gradually corrupted by noise, transforming it into a simpler, noise-dominated state. 
This forward noising process can be represented as follows:
\begin{equation}
q(x_T|x_0)=\prod_{t=1}^{T}q(x_t|x_{t-1}),
\end{equation}
\begin{equation}
    q(x_t|x_0) = \mathcal{N}(x_t;\sqrt{\overline{\alpha}_t}x_0,(1-\overline{\alpha}_t)I),
\end{equation}
where $\{x_t\}_{t=1}^{T}$ denotes a sequence of noised samples from time $t=1$ to $t=T$.
Then, DDPM learns the reverse process that recovers the original sample utilizing learned $\mu_\theta$ and $\Sigma_\theta$:
\begin{equation}
    p_\theta(x_{t-1}|x_t) = \mathcal{N}(x_{t-1};\mu_\theta(x_t),\sigma_\theta(x_t)),
\end{equation}
where $\theta$ represent the parameters of the denoiser, and are trained to minimize the mean squared error between $\epsilon_\theta(x_t)$ and the true Gaussian noise $\epsilon_t$:
$\min_\theta \|\epsilon_\theta(x_t)-\epsilon_t\|_2^2$.
Once $p_\theta$ is trained, new samples can be sampled by initializing $x_t\sim\mathcal{N}(0,\mathbf{I)}$, and sampling $x_{t-1}\sim p_\theta(x_{t-1}|x_t)$ via the reparameterization trick.

\subsubsection{Architecture}
We propose a novel architecture, the U-Net-style Diffusion Transformer (UDiT), which combines the U-Net structure with the Diffusion Transformer, as shown in Figure \ref{udit}.
After corrupting the samples in the training set with varying levels of Gaussian noise and vectorizing to fixed-length sequences, the input tokens are processed by a series of Diffusion Transformer layers composed of multiple DiT blocks.
In addition to noised inputs, the diffusion model meanwhile handles additional conditional information, such as the noise timesteps $t$, class labels $c$, natural language, etc.

Drawing inspiration from the original U-Net design, UDiT features an ``encoder-decoder'' architecture. 
The encoder compresses sequence length through downsampling operations, 
while the decoder restores the sequence to its original size via upsampling.
We designate five distinct stages in the process, illustrated in Figure \ref{udit}b: the sequence is downsampled twice during encoding to reduce the sequence length, and in the decoding stage, the sequence is upsampled to restore its original size. 
At each stage, each DiT layer is composed of multiple DiT blocks (Figure \ref{udit}c), with the block size changing as the downsampling and upsampling operations across stages.
Inspired by the Residual U-ViT designs \cite{hoogeboom2023simple, hoogeboom2024simpler}, 
we remove the block-wise skip-connections and instead employ residual connections (RC) between different layers, as shown in Figure \ref{udit}d.
Furthermore, we adopt an asymmetric network design, varying the number of DiT blocks across different stages.

After processing with the final DiT block, we need to decode the sequence-form tokens into the corresponding noise predictions. 
This decoding is implemented using a standard linear decoder combined with a final layer norm. 
Similar to the DiT block design, parameter regression is employed in the final layer. 
Finally, the decoded tokens are rearranged to map back to their original spatial layout.

The following provides detailed explanations of each component in the UDiT architecture.
\paragraph{DiT Block.}
Based on DiT's exploration of different block designs, we adopt the Adaptive layer norm-Zero (\textit{adaLN-Zero}) block as the core component of our architecture (Figure \ref{udit}c).
Each block comprises a self-attention module as the token mixer and a feed-forward network as the channel mixer.
Temporal and conditional embeddings, derived from the time step $t$ and condition $c$, are combined via summation to regress the scaling $\gamma$ and shift $\beta$ parameters for adaptive layer normalization. 
Additionally, scaling factors $\alpha$ are regressed and applied to the input dimensions prior to residual connections, providing finer control over feature transformations.


\paragraph{Residual Connection.}
In traditional U-Net architectures, skip-connections are typically used to bypass sub-sampled levels, such as those involving average pooling. 
However, this approach diverges significantly from the structure of large language models, which rely solely on residual connections without incorporating skip connections.
To streamline the model structure, we replace conventional skip-connections with residual connections \cite{hoogeboom2024simpler}, as shown in Figure \ref{udit}d. 
At each corresponding subsampling and upsampling operation, we introduce a simplified residual form of skip-connection, which effectively replaces the more complex traditional skip-connections for each stage.

To formalize the structure, let $f_d$ and $f_u$ represent the downsampling and upsampling stages of UDiT, respectively, and let $f_m$ denote the middle stage. 
The two-stage residual connection for UDiT can be expressed as follows:
\begin{equation}
    f(x) = f_u\left(U\left(f_m(D(h))-D(h)\right)+h\right),
\end{equation}
where $h = f_d(x)$ and $D$ and $U$ correspond to the downsampling and upsampling operations, respectively.
In this paper, we utilize a convolutional layer for downsampling to reduce the feature dimensionality, while upsampling is achieved using interpolation followed by a convolutional layer to restore the input feature dimensionality. 
Furthermore, in multi-stage architectures, the middle stage $f_m$ can itself recursively incorporate additional downsampling and upsampling stages, allowing the definition of hierarchical feature transformations across multiple levels of the UDiT architecture.


\paragraph{Assymetric Design.}
In the proposed UDiT model, each stage incorporates a single skip-connection, allowing for the flexible use of varying numbers of DiT blocks at different stages. 
Leveraging the advantages of a decoder-only architecture, we shift computational focus from the encoder stage (the downsampling half of the model) to the decoder stage (the upsampling half of the model), resulting in a higher number of DiT blocks in the decoder. 
This design choice aligns with findings in prior studies \cite{vaswani2017attention, hoogeboom2023simple}, which demonstrate that decoder-dominated architectures tend to exhibit superior performance in generative tasks.

Building upon this foundation, given that the sequence lengths vary across different stages, the temporal and class embedding dimensions in the DiT blocks can be set to fixed sizes.
In addition, the feature dimensions and the number of attention heads can be customized for each stage. 
Specifically, intermediate layers can utilize higher feature dimensions and fewer attention heads to enhance precision, thereby balancing computational efficiency and model accuracy across the architecture.



\begin{figure*}[ht]
\centering
\centerline{\includegraphics[width=0.98\textwidth]{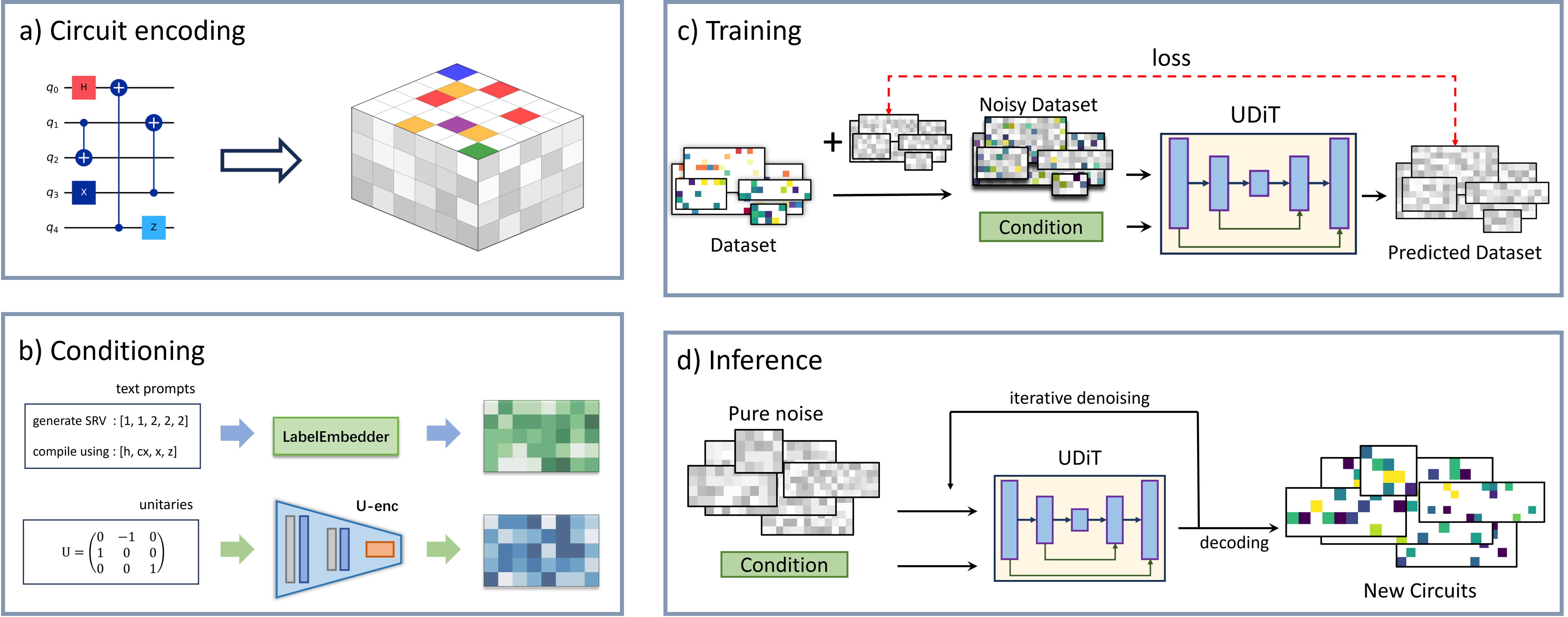}}
\caption{\textbf{The pipeline of the UDiTQC framework: (a)} Quantum circuit are encoded as two-dimentional tensors, then embedding into high-dimensional continuous tensor space. \textbf{(b)} Conditional embedding in UDiTQC, with SRV or compilation gate sets embedded through class labels; unitary encoder is trained with UDiT to create the encoding of an input unitary. \textbf{(c-d)} Schematic representation of the diffusion model training process and posterior inference from the trained model. }
\label{pipeline}
\vspace{-0.4cm}
\end{figure*}

\subsection{The Proposed UDiTQC}
By integrating quantum circuit encoding-decoding methods and specific conditioning embeddings, we propose UDiTQC, a framework that uses the UDiT-based diffusion model to generate quantum circuits. The pipeline for UDiTQC is illustrated in Figure \ref{pipeline}.

To enhance the efficiency and representational capacity of UDiTQC, we adopt a tensor-based representation for quantum circuits inspired by \cite{fosel2021quantum}.
In this formulation, each quantum circuit is represented as a two-dimensional tensor, where the first dimension encodes the qubit indices, and the second dimension encodes the time steps for gate placement, constrained to one gate per time step as in Figure \ref{pipeline}a.
We fix an arbitrary-length representation prior to training, assigning each gate a randomly generated orthogonal high-dimensional continuous embedding.
Using a vectorization strategy similar to the \textit{patchify} mechanism in DiT \cite{peebles2023scalable}, we transform the tensor into a fixed-length sequence while simultaneously performing feature expansion. Next, we apply frequency-based positional embeddings (sine-cosine formulation) to encode temporal and spatial relationships within the circuit, enabling the transformer layers to model both local and global dependencies.
During the decoding process, the generated quantum circuits are mapped to the predefined gate set by computing the cosine similarity between the described gate and the predefined gate embeddings, ensuring that the generated circuits align with the predefined gate set and comply with the physical constraints of quantum devices (see Appendix \ref{qc encoding} for details).

In Figure \ref{pipeline}b, we encode the requirements of the circuit as conditions, such as the SRV of the circuits in entanglement generation or the gate set used in unitary compilation tasks. These discrete and finite forms can be converted into labels, which are embedded into the UDiT model through a LabelEmbedder. For unitary embeddings, we use a U-enc, which is trained concurrently with the model, to encode the unitaries and combine it with the label before inputting it into the DiT block, as detailed in the Appendix \ref{cond}.

To train the diffusion model, we corrupt the samples in the training set (encoded circuits) with varying levels of Gaussian noise (Figure \ref{pipeline}c). The noisy training set, along with the associated conditions, is then input into the UDiT model to learn the noise present in the samples. The model learns to predict and subtract the noise from the input samples, effectively denoising them. After training, a fully noisy tensor is fed into the model, which iteratively denoises it under the guidance of the chosen condition using rescaled classifier-free guidance \cite{ho2022classifier}, and generates high-quality samples after several steps (Figure \ref{pipeline}d). Details regarding training and inference are provided in Appendix \ref{TaI}.

\section{Experiments}
We first evaluate the compatibility of the UDiT model through an ablation study, detailed in Appendix \ref{ablation}.
This study assesses various model designs based on 3-qubit entanglement generation tasks. The configurations tested include modifications to the architecture such as sequence-based gate embeddings, U-Net-style modifications, asymmetric structures, residual connections, and adjustments to the feature dimensions during downsampling and upsampling. The results from these experiments are summarized in Table \ref{tab:ablative_results}, where we compare the training speed, average accuracy, and entanglement generation accuracy across different designs.

Having validated the effectiveness of the UDiT architecture, we then apply it to two distinct quantum circuit generation problems: entanglement generation and unitary compilation. 
The entanglement generation task serves as a benchmark for evaluating the model's potential across different scenarios, while the unitary compilation task represents a fundamental challenge in the field.
In both tasks, the model is trained using the same denoising loss function, with different tasks achieved by modifying the training samples and their conditioning. The primary difference between the tasks lies in the labeling scheme, where task-specific label categorizations are applied, though the underlying setup remains consistent.

\begin{figure*}[ht]
\centering
\centerline{\includegraphics[width=0.85\textwidth]{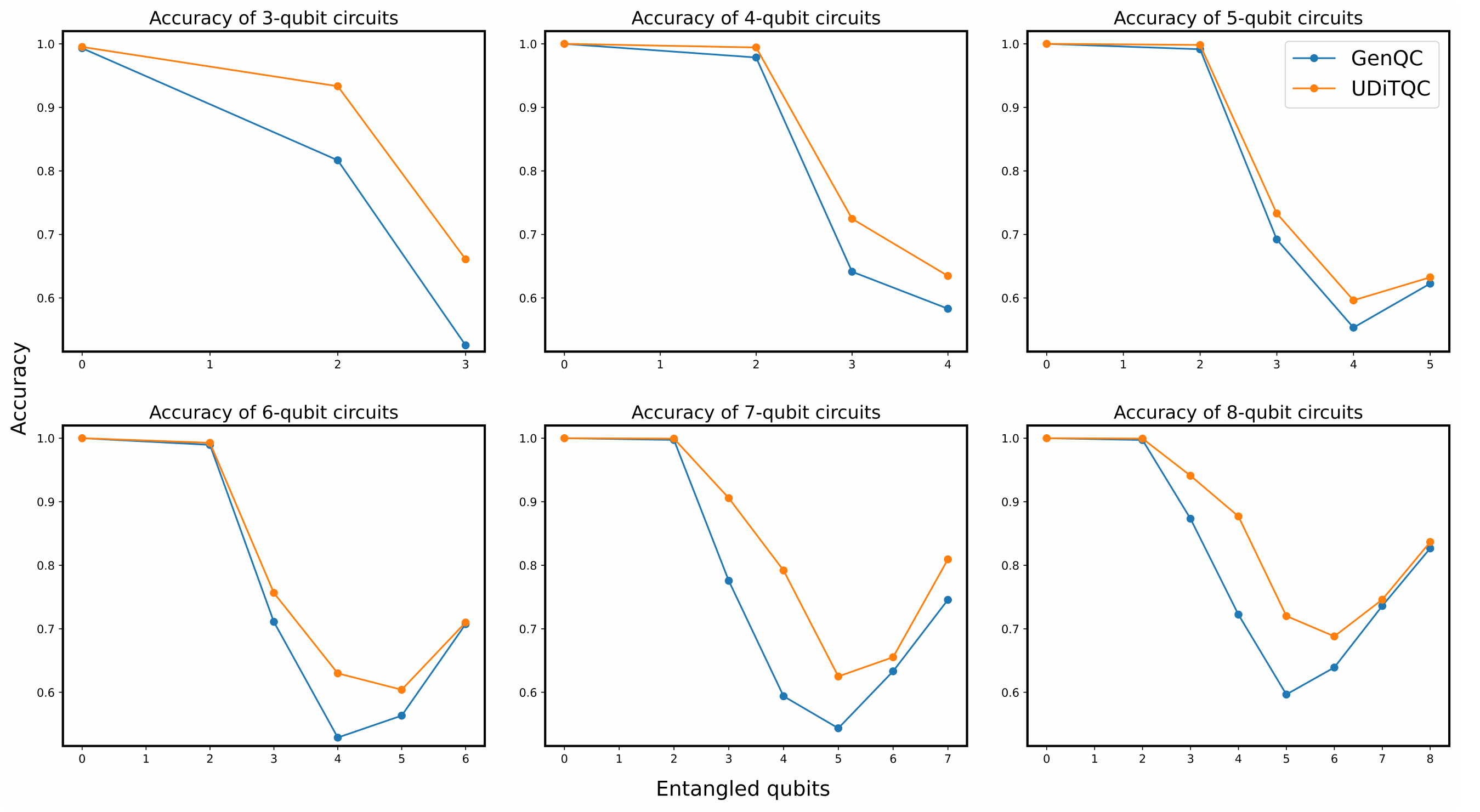}}
\caption{Model accuracy vs. the number of entangled qubits for circuits with different numbers of qubits. The accuracy represents the average generation accuracy for SRVs corresponding to each number of entangled qubits.}
\label{srv acc}
\vskip -0.2in
\end{figure*}

\subsection{Entanglement Generation}

Our objective is to generate quantum circuits with predefined entanglement states, characterized by their Schmidt Rank Vector (SRV) \cite{huber2013structure}, 
which provides a quantitative representation of entanglement structure through a numerical vector that indicates the rank of reduced density matrices for each subsystem.

To evaluate the efficacy of our proposed method, we conducted extensive experiments focusing on quantum circuit generation with specific entanglement states.
Our experimental framework encompassed circuits of varying qubit counts, analyzing their entanglement properties, and assessing the method's generalizability across different circuit configurations and lengths (with detail presented in Appendix \ref{ap:dataset}).

Following dataset balancing, we trained the UDiTQC model and conducted comprehensive experiments across various qubit configurations. 
For performance comparison, we replicated the GenQC experimental setup described in the literature \cite{furrutter2024quantum} using its corresponding code repository\footnote{\url{https://github.com/FlorianFuerrutter/genQC}} and trained with new datasets consistent with UDiTQC’s training setup.
The total number of circuits for each qubit count was approximately the same as reported, ensuring a fair and comprehensive comparison. 
During model training, the SRV of each circuit served as its class label, and all other training parameters were kept consistent across models. 
Upon completing training, we evaluated the model by generating circuits corresponding to various SRVs.
The model evaluation involved generating circuits corresponding to diverse SRVs, with all possible SRV configurations serving as generation prompts.
Performance comparison was based on the average precision of entanglement across identical qubit configurations.
\begin{figure}[ht]
\centering
\centerline{\includegraphics[width=0.85\columnwidth]{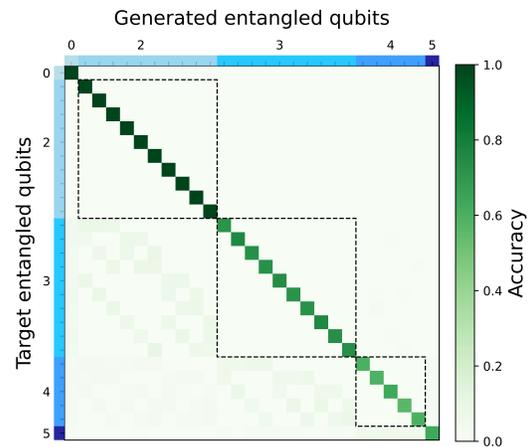}}
\caption{Confusion matrix comparing the input and generated SRVs for 5-qubit circuits. The SRVs are grouped by the number of entangled qubits for clarity.}
\label{confusion matrix}
\vspace{-0.4cm}
\end{figure}


As demonstrated in Figure \ref{srv acc}, UDiTQC consistently achieved superior generation accuracy compared to GenQC across experiments involving 3 to 8 qubits. 
Figure \ref{confusion matrix} provides a detailed accuracy analysis for 5-qubit circuits, revealing successful entanglement state generation in the majority of cases, though accuracy exhibits a modest decline with increasing numbers of entangled qubits.
This degradation can be attributed to the heightened complexity in both design and execution of quantum circuits requiring more gates to achieve higher degrees of entanglement.
Significantly, UDiTQC demonstrates robust generalization capabilities, successfully generating novel circuits beyond the training set while maintaining diversity and adherence to target SRV constraints. 
This capacity for producing unique, previously unseen quantum circuits while preserving predetermined entanglement characteristics underscores the model's versatility and potential impact.

UDiTQC demonstrates superior performance in generation accuracy, scalability, and efficient learning of gate placement patterns, which validates UDiTQC's effectiveness and generalizability in quantum circuit design. 
The model consistently respects predefined constraints, including gate set specifications and one-gate-per-time-step requirements.

\subsection{Masking and Editing Circuits}

\begin{figure}[!t]
    \centering
\includegraphics[width=\columnwidth]{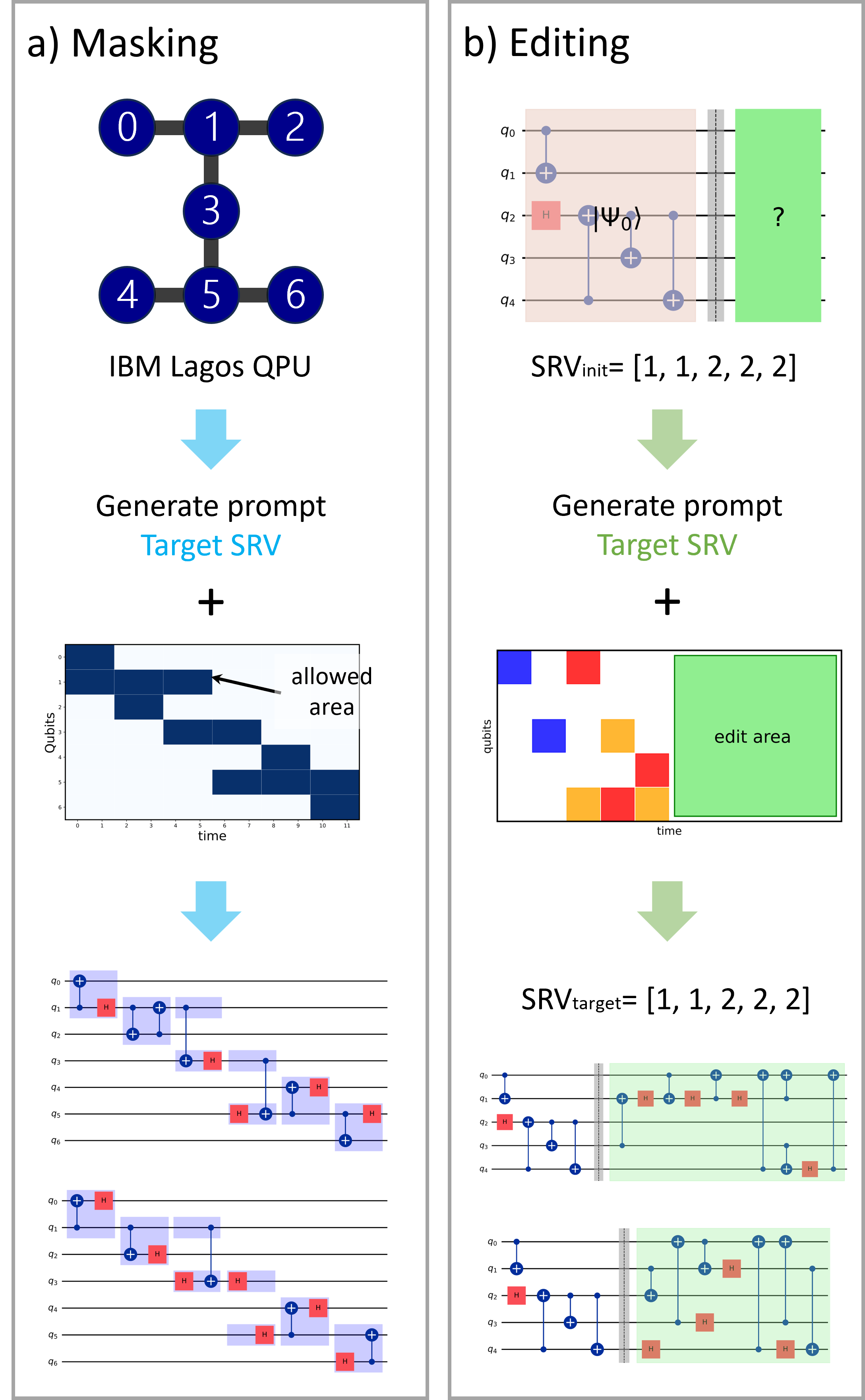}
    \caption{\textbf{Masking and editing circuits. (a)} Masking: The layout of a quantum processor can be embedded as a mask, preventing the model from placing gates at specific locations in the input tensor (represented by the white area). \textbf{(b)} Editing: Portions of the circuit can be fixed to specific gates before generation, such as to incorporate an initial quantum state, ensuring that the desired quantum computation is performed on this state.}
\label{masking and editing}
\vskip -0.2in
\end{figure}

Building upon entanglement generation, the quantum circuits corresponding to the tensors can be edited or masked, as proposed for image editing \cite{lugmayr2022repaint}.
These operations enable two distinct functionalities: one prevents the model from placing gates in certain areas of the circuit, and the other imposes the presence of specific gates.
These operations do not require any additional training, but rely solely on adding different conditions during inference on the entanglement generation model.

\paragraph{Masking.}
The masking operation serves as a fundamental mechanism for imposing specific structural constraints on circuit generation. 
When operations must be performed between spatially distant qubits, they necessitate decomposition into a series of intermediate gates and exchange operations throughout the circuit, resulting in diminished gate fidelity \cite{baumer2024efficient}. These physical limitations, while crucial for practical implementation, are not inherently represented in conventional circuit diagrams, which suggest unrestricted qubit connectivity.

As illustrated in Figure \ref{masking and editing}a, we enforce specific physical constraints by masking designated sections of the model input tensor, effectively preventing gate placement within these masked regions. 
The masked tensor is then fed into the model along with the corresponding SRV label to generate the appropriate circuits. 
The masked tensor, in conjunction with the corresponding SRV label, is then processed by the model to generate circuits that comply with the specified constraints. 
This approach offers flexible control over fundamental circuit parameters, including qubit count and circuit length, through techniques such as complete row masking for specific qubits.
The efficacy of this constraint enforcement mechanism is demonstrated in the lower portion of Figure \ref{masking and editing}a, where the generated quantum circuits explicitly satisfy the imposed physical limitations. While these constrained circuits may exhibit marginally reduced accuracy compared to unconstrained generation tasks, the model successfully maintains adherence to the specified physical constraints. This capability significantly enhances the practical applicability of our approach across diverse quantum computing scenarios.


\paragraph{Editing.}
The editing operation facilitates the precise specification of predetermined gates throughout the circuit prior to initiating the diffusion process. 
This capability addresses practical scenarios where a quantum state of interest, represented by an initial circuit configuration, serves as the foundation for subsequent computational operations. 
As illustrated in the Figure \ref{masking and editing}b, this process begins with an input circuit comprising five gates in the initial state, with the objective of transforming it to achieve a target SRV while preserving the initial five-gate configuration.
\begin{figure}[t]
    \centering
    \includegraphics[width=0.9\columnwidth]{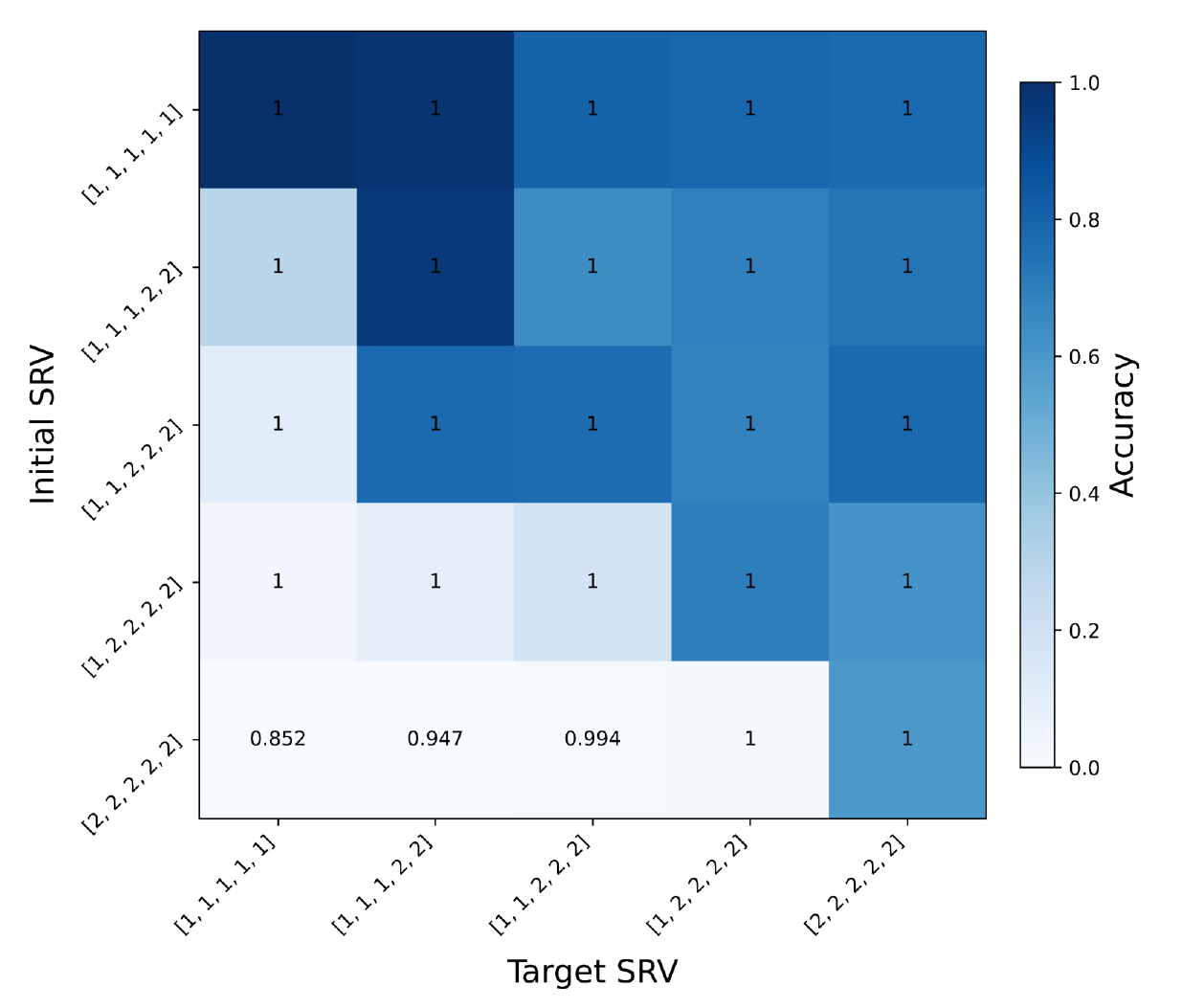}
    \caption{Accuracy of editing circuits from an input SRV to a target SRV. The numbers indicate the percentage of initial circuits for which at least one solution was found within a sample of 1024 generated circuits.}
\label{editting acc}
\vspace{-0.4cm}
\end{figure}

The experimental results demonstrate successful constraint application in generating circuits that achieve the target SRV specifications (Figure \ref{editting acc}). 
Notably, the method exhibits robust performance in tasks requiring increased entanglement (elements above the diagonal). However, significant challenges emerge in entanglement-reducing operations (elements below the diagonal). The reduced accuracy in these cases can be attributed to the complexity of the requisite gate sequence: the process necessitates applying multiple gates to the initial high-entanglement state before implementing the editing operations to achieve the desired low-entanglement configuration.
Despite these inherent challenges in entanglement reduction, the model maintains commendable performance, successfully executing transitions from high-entanglement to low-entanglement states with an efficacy rate exceeding $85.2\%$. This achievement underscores the model's robustness in handling complex quantum state transformations, even under demanding constraints that require substantial modifications to the circuit's entanglement properties.

\subsection{Unitary Compilation}

We train UDiTQC for a critical task in quantum circuit generation—compiling unitaries into circuits. 
We generate a variety of 3-qubit circuits of different lengths and gate sets, and compute their corresponding unitary matrices to construct the dataset.
The conditions of the gate sets were used as the model's class labels, while the unitary compilation was integrated with an encoder trained jointly with the model.

\begin{figure}[!t]
\centering
\centerline{\includegraphics[width=\columnwidth]{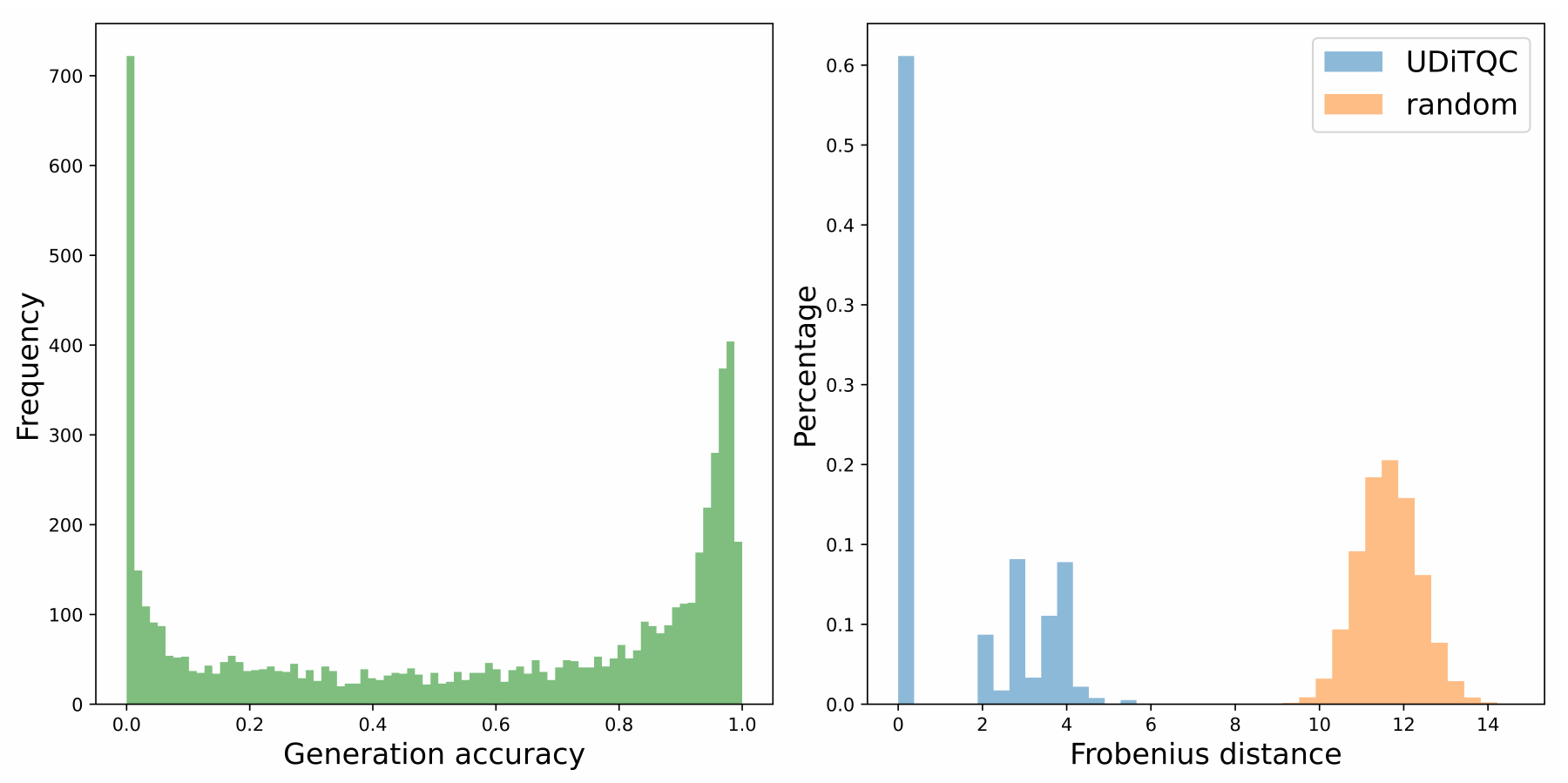}}
\caption{\textbf{Result of unitary compilation}. \textit{Left:} The frequency of generation accuracy, over 1024 generated circuits, for more than 5000 input unitaries. \textit{Right:} Frobenius distance between the target unitaries and circuits generated randomly or from random unitaries.}
\label{unitary}
\vspace{-0.4cm}
\end{figure}

After training, new unitaries were input into the model to evaluate its compiling capability.
During testing, we selected over five thousand unitary matrices that are not within the training set and compiled into 1024 circuits. 
Remarkably, UDiTQC achieved a compilation accuracy of $94.9\%$, outperforming GenQC, which achieved $92.6\%$, and the accuracy of the circuits generated for each unitary was consistently higher than that of GenQC. As shown in the Figure \ref{unitary}, the model was able to compile multiple distinct circuits for most unitary matrix instances, allowing prospective users to choose the most suitable circuit based on their needs.
Further, we assessed the model's ability by calculating the Frobenius norm $\frac{1}{2}\|U_t-U_g\|_F^2$ of the distance between the generated circuit's unitary $U_g$ and the target unitary $U_t$. 
As shown in the right-hand figure, the majority of target unitary matrices were compiled with a Frobenius norm of zero, indicating perfect accuracy.
Even in cases where the norm was nonzero, the distance was significantly smaller compared to that of randomly generated unitary matrices, demonstrating the model's precision in unitary compilation.

\section{Conclusion}

In this paper, we propose UDiT, a novel architecture that fundamentally reimagines the DiT framework through a U-Net-style design, incorporating residual connections and asymmetric structures to enhance the capabilities of diffusion models. 
This architecture successfully combines U-Net’s multi-scale feature extraction with the global modeling advantages of Transformer. 
We further apply this architecture to quantum circuit synthesis, introducing UDiTQC, which leads to significant improvements in circuit conditioning and compilation generation.
Through comprehensive evaluation, UDiTQC demonstrates superior performance in both entanglement generation and unitary compilation tasks, while also supporting advanced circuit manipulation capabilities including masking and editing, achieving higher accuracy and efficiency compared to the state-of-the-art GenQC method.

The success of our framework extends beyond quantum applications, offering promising directions for fields where both local feature extraction and global context modeling are crucial. 
Future research will focus on expanding the UDiT model to a wider range of applications and exploring the transition from gate-based to measurement-based quantum computing paradigms. This dual contribution - which advances both quantum circuit synthesis and generative model architectures - marks a significant step forward at the intersection of quantum computing and machine learning research.

\newpage
\section*{Impact Statement}

This work advances machine learning techniques for quantum computing, improving the generation and compilation of quantum circuits. 
The methods developed have the potential to enhance quantum algorithm efficiency, with future applications in cryptography, optimization, and material science, driving progress in quantum technologies. 
While there are many potential societal consequences of our work, we feel none need to be specifically highlighted here at this stage.


\bibliography{reference}
\bibliographystyle{icml2025}

\newpage
\appendix
\onecolumn
\section{Quantum Circuit Encoding}
\label{qc encoding}
\begin{figure}[h]
    \centering
    \includegraphics[width=0.9\linewidth]{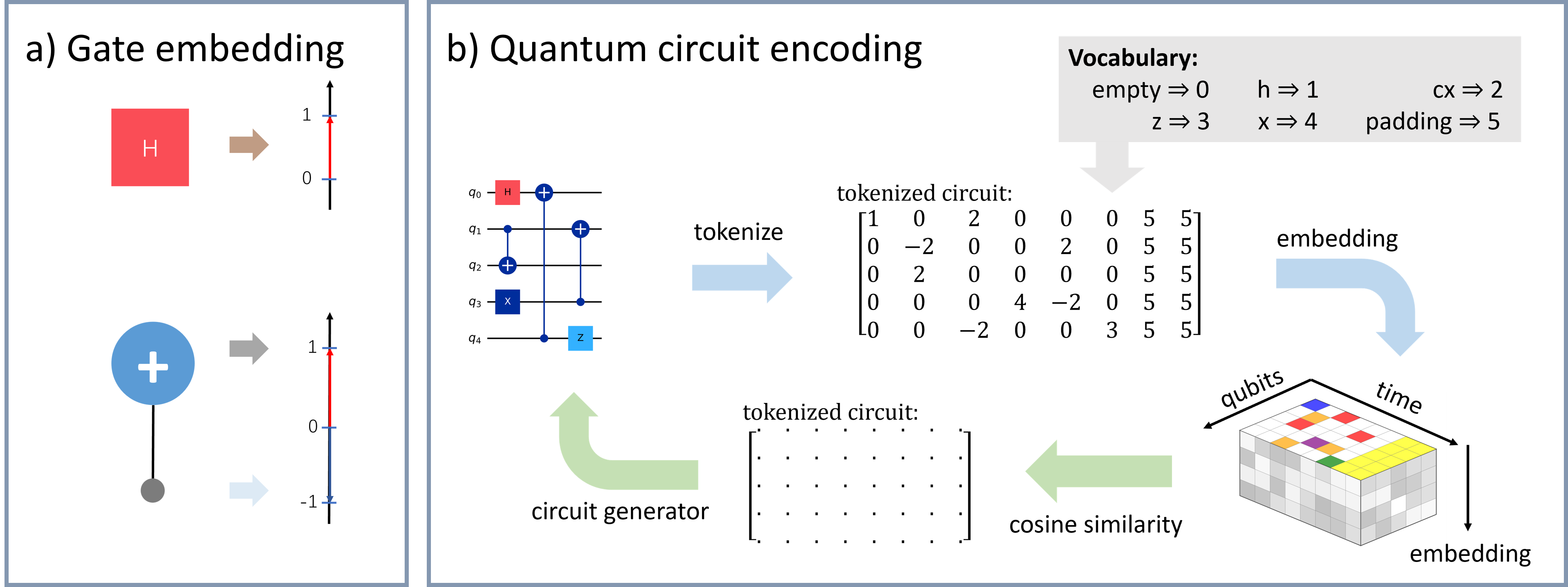}
    \caption{\textbf{Quantum circuit encoding.(a)} Schematic representation of the gate embeddings for a single and multi qubit gate. \textbf{(b)} The pipeline for encoding and decoding quantum circuits.}
    \label{circuit encoding}
\end{figure}

In this section, we present the procedure for encoding quantum circuits into continuous tensors, which can subsequently be fed into neural networks for processing.
we first address the embedding of quantum gates, as illustrated in the Figure \ref{circuit encoding}a.
For single-qubit gates, the corresponding embedding is represented by the numerical value associated with the qubit. For multi-qubit gates, the control and target nodes are described using the same numerical embedding, but with opposite signs. 
This approach ensures that both the control and target nodes of multi-qubit gates are captured in the embedding, reflecting their respective roles within the gate.

Building upon this, the quantum circuit is embedded into a two-dimensional tensor, where the first dimension corresponds to the qubits and the second dimension corresponds to time (i.e., the sequence of gates).
This tensor can be expanded to accommodate any time length (the number of gates) and any spatial size (the number of qubits).
The flexibility in the size of the tensor ensures that the embedding can represent quantum circuits with arbitrary complexity, while preserving the integrity of the circuit’s structure.

This encoding method guarantees both the completeness and flexibility of the quantum circuit embedding, allowing it to be applied across a wide range of circuit configurations. As a result, the proposed encoding approach is not limited to quantum computing but can be extended to other applications that involve time and space structured data.

The encoding and decoding pipeline is illustrated in Figure \ref{circuit encoding}b, with the blue arrows representing the encoding process and the green arrows representing the decoding process. 
The encoding process is divided into two parts. First, the quantum circuit is tokenized using gate embeddings, which transform the circuit into a two-dimensional integer matrix representation. 
This matrix is then further embedded into a normalized continuous input, centered around zero with a range of $[-1,1]$.
To ensure robustness in decoding, orthogonal embeddings are typically used for the gate embeddings. 
The feature dimension of the embedding is selected as $d = N+2$, where N is the number of gates.
The additional two dimensions correspond to padding and background, representing the fill in the unified circuit matrix and the zero information for positions without gates, respectively. The embedding for the circuit portion is usually determined prior to the training of the diffusion model.

In the decoding process, the first step involves converting the continuous tensor generated by the model back into a tokenized form.
We select the token k corresponding to the embedding $\mathbf{v}_k$ that has the highest cosine similarity with $\mathbf{v}_{gen}$:
\begin{equation}
    \tilde{k}= \mathop{\arg\max}_k|S_C(\mathbf{v}_k,\mathbf{v}_{gen})|.
\end{equation}
To resolve the node types, the second step involves using the formula below:
\begin{equation}
    k = \tilde{k}\cdot \text{sign}\ S_C(\mathbf{v}_{\tilde{k}},\mathbf{v}_{gen}).
\end{equation}
This calculation is performed for each spatiotemporal position in the tensor encoding, returning a tokenized integer matrix that precisely describes a valid circuit. Error circuits refer to cases where the model places two or more gates at the same time step or fails to properly position the target and control nodes for multi-qubit gates. 
Interestingly, we observe that the trained model matches the original embeddings with high accuracy (i.e., $\mathbf{v}_k\approx\mathbf{v}_{gen}$), and we assume that it is possible to "overload" the vector space with ${R}^d$ with $N>d$ gates. 

Building upon this, the input is passed into UDiT, where the ``patchify'' operation is applied. The gate and time dimensions are embedded into a sequence of length K ($K = max\_qubits \times max\_gates$). This step ensures that the quantum circuit is represented as a fixed-length sequence, making it suitable for input into UDiT. The embedding process captures both gate-specific and time-specific information, preserving the structural properties of the circuit.

\section{Conditioning}
\label{cond}
In this section, we provide a detailed description of the conditional encoding methods used in quantum circuits. Since UDiT is a conditional diffusion model, it is necessary to embed the corresponding labels or features of the circuits into the model during training to guide the diffusion model's learning process.
\subsection{Label Embedding}
For the entanglement generation task, the goal is to generate circuits that produce specific types of entanglement given the circuit size. The criterion for entanglement is the SRV (Schmidt Rank Vector) of the circuit. 
Therefore, in this task, the SRV serves as the label for the circuits in the dataset. 
Since the SRV is represented as a numerical vector, where each position indicates whether the entanglement state is 1 or 2, the circuits corresponding to a fixed set of qubits have a finite number of representations. These representations can be numbered, allowing the label in the data to be transformed into a class label.

For the unitary compilation task, different subsets of compilation gates correspond to different circuits. Therefore, the selected gate subset can serve as the class label for this task. For a given gate set, we can generate corresponding sub-sets in lexicographical order, which are then mapped to integer labels.

Once these labels are converted into class labels, they are embedded into continuous vector spaces using a LabelEmbedder. Additionally, label dropout is employed to simulate classifier-free guidance during training.

\subsection{Unitary Embedding}
The second part focuses on the embedding of unitary matrices in the unitary compilation task, referred to as \textbf{U-enc}.
The U-enc implementation starts by creating an input tensor from the unitary matrix. The real and imaginary parts of the matrix are separated into two channels, which are then fed into the unitary encoder. After an initial convolutional layer, a 2D positional encoding layer is introduced to encode the absolute positions of the unitary elements. A Transformer encoder with self-attention layers is employed to capture global dependencies. This global attention mechanism is crucial for handling unitary matrices, as the information they contain is non-local. Between the attention blocks, a downsampling layer with a kernel size of 2x2 is introduced, followed by convolutional layers that expand the unitary condition into the corresponding hidden dimensions.

This approach ensures that the unitary condition is efficiently processed and integrated into the model, enabling effective learning of unitary transformations in the compilation task.

\section{Training and Inference}
\label{TaI}
\subsection{Training}

We train the UDiT model based on the denoising diffusion probabilistic models (DDPM) framework.
Specifically, we parameterize the model as an $\epsilon$-predictor. At each time step $t$, the model learns to predict the noise term $\mathbf{\epsilon}_t$ of the noisy tensor $\mathbf{x}_t = \sqrt{\bar{\alpha_t}}\mathbf{x}_0 + \sqrt{1 - \bar{\alpha_t}}\mathbf{\epsilon}_t$, where $\mathbf{\epsilon}_t \sim \mathcal{N}(0, \mathbf{I})$, $\mathbf{x}_0$ is a sample from the training dataset with distribution $q(\mathbf{x}_0)$, and $\sqrt{\bar{\alpha_t}}$ denotes the variance schedule. We define the cumulative variance as $\bar{\alpha}_t = \prod_{i=0}^{t} \alpha_i = \prod_{i=0}^{t} (1 - \beta_i)$.

The denoising model, with parameters $\theta$, predicts the noise $\mathbf{\epsilon}_\theta$ conditioned on an additional context $\mathbf{c}$. The model parameters are optimized to minimize the following loss function:
\begin{equation}
    \mathcal{L}=\mathbb{E}_{t \sim \mathcal{U}[0, T], \mathbf{x}_0 \sim q\left(\mathbf{x}_0\right), \boldsymbol{\epsilon}_t \sim \mathcal{N}(0, \mathbf{I})}\left[\left\|\boldsymbol{\epsilon}_t-\boldsymbol{\epsilon}_\theta\left(\mathbf{x}_t, t, \mathbf{c}\right)\right\|_2^2\right],
\end{equation}
where the expectation is taken over the time step $t$, the clean data $\mathbf{x}_0$, and the noise term $\boldsymbol{\epsilon}_t$.

For the variance schedule, we use the squared cosine $\beta$ schedule as proposed by \cite{nichol2021improved}. We train the model for $T = 1000$ diffusion steps using the AdamW optimizer \cite{loshchilov2017decoupled} with a one-cycle learning rate policy \cite{smith2019super}. The initial learning rate is set to $3 \cdot 10^{-4}$, and we train for a total of 300 epochs.

During training, both the time and class label embeddings in UDiT are jointly trained. These discrete time or class labels are embedded into continuous tensors, which are then used as conditions for the DiT block. For the unitary compilation task, the encoder for the unitary matrix is also trained concurrently, treating the corresponding part of the condition as $c_\Phi(U)$, where $U$ represents the given unitary matrix and $\Phi$ denotes the parameters of the encoder.

In our dataset, due to the extensive random sampling in the unitary space, most unitaries have a unique circuit implementation. This can lead to overfitting, as the model may memorize the one-to-one mapping between the unitary matrices and their tensor encodings. To mitigate this issue, we apply dropout within the unitary encoder (U-enc). 


For the condition embedding in UDiT, the time and label embeddings are summed together to form the overall condition, which is then used for adaptive normalization in the DiT block. In the unitary compilation task, unitary information is incorporated by concatenating the unitary matrix embedding with the time and label embeddings, followed by a linear layer to either expand or reduce the feature dimensions.

\subsection{Inference}
After training, we sample (infer) new circuits from the model across various scenarios. 
To sample a new circuit, we provide the model with a condition $\mathbf{c}$ and a noise input tensor $\mathbf{x}_T \sim \mathcal{N}(0, \mathbf{I})$. 
The noise tensor influences the size of the output circuit, specifically the number of qubits and the maximum number of gates that the model can place.

Since the model has been trained on circuits with various padding configurations, it can effectively place padding tokens when needed. Therefore, a reasonable strategy is to provide the model with a sufficiently long tensor and allow it to limit the total number of gates by placing padding tokens as necessary.
For the sampling method, we employ the DDPM sampler with a reduced number of denoising steps, as introduced by the Denoising Diffusion Implicit Models (DDIM).


Conditional diffusion models incorporate additional information, such as a class label $c$, to guide the generative process. To sample according to the given condition, we employ rescaled classifier-free guidance (CFG) during inference.
In this framework, the reverse process is presented as $p_\theta(x_{t-1}|x_t,c)$, where both $\epsilon_\theta$ and $\Sigma_\theta$ are conditioned on $c$.
By interpreting the output of diffusion models as the score function, classifier-free guidance \cite{ho2022classifier} can be leveraged to direct the DDPM sampling procedure toward finding $x$ that maximize $\log p(c|x)$ by: 
\begin{equation}
    \hat{\epsilon}_\theta(x_t,c) = \epsilon_\theta(x_t,\emptyset) + s\cdot\nabla_x\log p(x|c)\varpropto \epsilon_\theta(x_t,\emptyset)+s\cdot(\epsilon_\theta(x_t,c)-\epsilon_\theta(x_t,\emptyset)),
\end{equation}

where $s>1$ indicates the scale of the guidance, with $s=1$ recovering the standard sampling process.
To evaluate the diffusion model under $c=\emptyset$, class labels $c$ are randomly dropped out during training and replaced with a learned "null" embedding $\emptyset$.
For all experiments, we select $cfg\_scale=7.5$.

For the entanglement generation task, including masking and editing, we generate 1024 new circuits for each SRV condition using 100 denoising steps in all tests to evaluate accuracy.
In unitary compilation inference, for each given unitary, we generate 1024 circuits with 50 denoising steps to compare the results.

We observe that the masking task is more challenging than the other tasks due to the strict constraints applied to the generated circuits. 
Specifically, we notice an increase in the number of erroneous circuits, with the error rate rising from less than $1\%$ in the non-masked entanglement generation task to approximately $80\%$ in this case.
Furthermore, the number of samples required to find an appropriate solution varies with the severity of the masking constraint. 
In contrast, the editing task is less stringent, only masking the first few gates of the quantum circuit. The model can still generate valid circuits, but the accuracy of the generated SRV corresponding to the target decreases compared to when no editing constraint is applied.

\section{Ablation Study}
\label{ablation}
\begin{table}[h!]
\centering
\begin{tabular}{c|c|c|c}
\hline
\textbf{Model Design} & \textbf{Training Speed} & \textbf{Average Accuracy} & \textbf{Entangled Accuracy} \\
\hline
DiT & 10.56 & 64.08 & 44.1 \\
DiTsq & 15.1 & 77.52 & 64.75 \\
U-Net-Style DiTsq & 10.74 & 79.18 & 69.2 \\
U-Net-Style DiT & 9.24 & 84.14 & 56.06 \\
+ Asymmetric Structure & 9.87 & 85.1 & 60.54 \\
+ Residual Connections & 8.56 & 86.2 & 61.97 \\
+ Hidden Feature Expansion & 9.9 & 82.4 & 59.14 \\
+ Fixed Hidden Feature & \textbf{15.5} & \textbf{89.12} & \textbf{65.72} \\
\hline
\end{tabular}
\caption{\textbf{Ablation study results of various model designs:} Training speed refers to the number of steps processed per second on the same machine. Average accuracy represents the mean accuracy for generating 5 distinct SRV for 3-qubit circuits, while entangled accuracy evaluates the model's ability to generate fully entangled quantum states, which is considered a key criterion for 3-qubit circuits due to its relatively low precision compared to other SRVs.}
\vskip -0.2in
\label{tab:ablative_results}
\end{table}

In this section, we evaluate the contribution of individual components in UDiT, starting with a base experiment on 3-qubit entanglement generation. 
Table \ref{tab:ablative_results} presents the results of the ablation study, comparing the performance of various model configurations in terms of training speed, average accuracy, and the accuracy of generating fully entangled states.

We begin by testing the original DiT method and then explore variations in the embedding of the circuit tensors. 
In one variant, we embed the gate representations at each timestep as a sequence, called DiTsq. 
To further enhance the model, we apply a U-Net-Style approach using convolutional layers for downsampling and upsampling, which only alters the sequence length of the tensor. 
This modification improves the DiT format and integrates the sequential embedding method. However, the introduction of additional convolutional layers for feature transformation slightly reduces the runtime efficiency.

Although U-Net-Style DiTsq achieves high accuracy in generating fully entangled states, its average accuracy is lower than that of the U-Net-Style DiT, and the scalability of this embedding method is limited. 
Therefore, follow-up experiments focus only on the embedding strategies mentioned in this paper. 
Building on this, we introduce asymmetric UDiT and incorporate residual connections into the DiT layer. 
While this improves accuracy, it also results in a reduction in training speed. 
To further optimize the model, we modify the feature dimensions between DiT layers, expanding the features during downsampling and upsampling while reducing the sequence length. 
This approach lowers the computational complexity and accelerates training, particularly for longer sequences.

Finally, by fixing the conditions and input token dimensions and reducing the number of attention heads in the intermediate layers, we arrive at the final UDiT network architecture (The last row in the Table \ref{tab:ablative_results}). 
This configuration achieves the highest efficiency and accuracy in generation tasks.


\section{Datasets}
\label{ap:dataset}
In this section, we describe the dataset generation process for the tasks of entanglement generation and unitary compilation, as discussed in this work.

\subsection{Generation of Random Circuits}
For the entanglement generation task, we generate random circuits for a fixed number of qubits. 
First, we sample the number of gates to be placed in the circuit from a uniform distribution with a specified lower and upper bound. 
Then, we sample the gates to be placed in the circuit from a predefined gate pool. 
After constructing the circuit, we compute its corresponding SRV (Schmidt Rank Vector), which is converted into prompt such as "Generated SRV: [1,2,2]". These SRVs are then embedded as labels and fed into the UDiT model.

For the unitary compilation task, we generate all possible subsets of gates from the gate pool using lexicographical ordering, where these subsets correspond to the labels in UDiT. 
Based on the chosen gate subsets, circuits are generated randomly for a given number of qubits and gates, and the corresponding unitary is computed as an additional condition for UDiT.

\subsection{Circuit Optimization}
We employ \emph{Qiskit’s Qiskit.compiler.transpile} \cite{qiskit2024} function to optimize all the generated random circuits. 
This optimization step primarily removes consecutive redundant gates, thereby improving the efficiency of the circuits and enhancing the overall quality of the dataset. 
Training the model with this optimized dataset significantly improves its accuracy. After optimization, we also remove all duplicate circuits to further enhance the diversity and quality of the dataset.

\subsection{Dataset Balancing}
For the entanglement generation task, the random circuits generated by Qiskit exhibit a significant imbalance in the distribution of SRV types (e.g., circuits with SRVs composed solely of 1s or 2s appear disproportionately more frequently). 
Therefore, we balanced the dataset before training the UDiT model. 
This balancing process ensures that each SRV type is equally represented in the training set.

In the case of unitary compilation, we balance the dataset by adjusting the number of circuits corresponding to each gate subset. 
For most subsets, the number of circuits is balanced to the desired count. For certain subsets, where the number of non-duplicate circuits is smaller than the specified balance size, we retain all the circuits.

\subsection{Dataset Construction}
\begin{table}[ht]
\centering
\begin{tabular}{c|c|c|c|c|c|c|c}
\hline
\textbf{Task} & \textbf{Gate Pool} & \textbf{Qubits} & \textbf{Min Gates} & \textbf{Max Gates} & \textbf{Labels} & \textbf{Balanced Size} & \textbf{Total Circuits} \\
\hline
\multirow{6}{*}{\makecell{Entanglement \\ generation}} & \multirow{6}{*}{H, CX} & 3 & 2 & 16 & 5 & 40000 & 200000 \\
 & & 4 & 3 & 20 & 12 & 25000 & 300000 \\
 & & 5 & 4 & 28 & 27 & 17000 & 459000 \\
 & & 6 & 5 & 40 & 58 & 8100 & 469800 \\
 & & 7 & 6 & 52 & 121 & 4000 & 484000 \\
 & & 8 & 7 & 52 & 248 & 2100 & 520800 \\
\hline
{\makecell{Unitary \\ Compilation}} & \makecell{H, CX, Z, X, \\ CCX, SWAP} & 3 & 2 & 12 & 63 & 20000 & 925000 \\
\hline
\end{tabular}
\caption{\textbf{Training set parameters for tasks.} ``Gate Pool'' refers to the set of gates used in the task's circuits. In the entanglement generation task, ``Label'' refers to the number of possible SRV types of all circuits in this qubit, corresponding to $num = 2^{q}-q$, where $q$ represents the qubits. In unitary compilation, ``Label'' corresponds to the gates used for compilation.``Balanced size'' refers to the number of circuits corresponding to each class after balancing, and ``total circuits'' refers to the total number of circuits for this qubit.}
\label{tab:dataset_table}
\end{table}

Using the above steps, we generate datasets with the parameters specified in Table \ref{tab:dataset_table}. 
For the entanglement generation task, we use a multi-qubit datasets with circuits consisting of 3 to 8 qubits. 
For the unitary compilation task, we construct a dataset with over 920,000 distinct 3-qubit circuits, generated from over 300,000 different unitaries.

\end{document}